\documentclass[a4paper]{article}
\pdfoutput=1
\usepackage{color}
\definecolor{mypink1}{rgb}{0.858, 0.188, 0.478}

\usepackage[none]{hyphenat}
\usepackage{multirow}
 
 \newcommand{\dq}[1]{``#1''}

\newcommand{\eg}{\textit{e.g.,\ }}

\usepackage{enumitem}

\usepackage{INTERSPEECH2019}

\title{Predicting Behavior in Cancer-Afflicted Patient and\\Spouse Interactions using Speech and Language}

\name{Sandeep Nallan Chakravarthula$^{1*}$\thanks{* Authors have equal contribution.}, Haoqi Li$^{1*}$, Shao-Yen Tseng$^1$,\\ Maija Reblin$^2$, Panayiotis Georgiou$^1$}

\address{
  $^1$Dept. of Electrical and Computer Engineering, University of Southern California, USA\\
  $^2$Dept. of Health Outcomes and Behavior, Moffitt Cancer Center, USA}
\email{\{nallanch,haoqili,shaoyent\}@usc.edu, maija.reblin@moffitt.org, georgiou@sipi.usc.edu}

\begin{document}

\maketitle

\begin{abstract}
\looseness=-1
Cancer impacts the quality of life of those diagnosed as well as their spouse caregivers, in addition to potentially influencing their day-to-day behaviors.
There is evidence that effective communication between spouses can improve well-being related to cancer but it is difficult to efficiently evaluate the quality of daily life interactions using manual annotation frameworks.
Automated recognition of behaviors based on the interaction cues of speakers can help analyze interactions in such couples and identify behaviors which are beneficial for effective communication.
In this paper, we present and detail a dataset of dyadic interactions in 85 real-life cancer-afflicted couples and a set of observational behavior codes pertaining to interpersonal communication attributes.
We describe and employ neural network-based systems for classifying these behaviors based on turn-level acoustic and lexical speech patterns.
Furthermore, we investigate the effect of controlling for factors such as gender, patient/caregiver role and conversation content on behavior classification.
Analysis of our preliminary results indicates the challenges in this task due to the nature of the targeted behaviors and suggests that techniques incorporating contextual processing might be better suited to tackle this problem.

\end{abstract}

\section{Introduction}
\label{sec:intro}
In 2019, the American Cancer Society estimates that there will be 1,762,450 new cancer cases diagnosed and 606,880 cancer deaths in the United States; a recent national report conservatively estimated that at least 2.8 million Americans were providing care to an adult with a primary diagnosis of cancer \cite{cancer2019report,national2016caregiving}.
Cancer impacts both patient and caregiver, particularly when the cancer becomes life-limiting.
Reviews of the literature on couples coping with cancer have suggested that communication is an important process through which couples make sense of their experience and coordinate coping \cite{li2014literature, traa2015dyadic}.

Effective couples communication can facilitate decision-making, improve intimacy and relationship quality, and improve well-being for both patient and caregiver.
However, evidence is scant when it comes to determining what might be the normal or even ideal amount or content of communication about cancer for patients and their spouse caregivers \cite{badr2017new}.
As such, more research is needed to describe how patients and their spouse caregivers communicate.
A major barrier towards interpersonal communication research to date is the inability to process communication data efficiently and effectively given the reliance on human coders \cite{reblin2018everyday}.

These challenges hint at the potential benefits of involving automatic behavior annotation, in which data-driven machine learning techniques are employed to automatically extract behavioral information directly from data, rather than relying on time-consuming and expensive annotations from human experts. 
Such behavior analysis work has been shown to be effective at identifying behaviors during interactions in domains such as couple therapy \cite{black2013toward, li2016_sparsely-connec, tseng2016couples}, depression \cite{gupta2014multimodal,nasir2016multimodal,tanaka2017brain} and suicide risk assessment \cite{cummins2015review, venek2017adolescent,baucom2017promise,nasir2017complexity}. 
However, due to potential domain mismatch, obtaining accurate performance in one domain by utilizing well-trained behavior analysis systems from a different domain is not straightforward.
A targeted effort at curating the appropriate data and domain-specific behavior coding is required in order to ascertain the feasibility and efficacy of such techniques for a new domain.

Towards this direction, in this paper, we present a dataset of audio recordings, transcriptions and behavior annotations of dyadic interactions between 85 real-life cancer-afflicted couples. 
We describe a deep learning-based system to classify the emotional behavior of spouses directly from their acoustic and lexical speech patterns.
We also present preliminary results on the effect of controlling for factors such as gender, role (patient/caregiver) and conversation content on the classification performance.

This paper is structured as follows: 
Sec.~\ref{sec:data} details the procedure used to collect the couples interaction dataset and provides a summary of the behavior codes used to annotate them.
Following this, in Sec.~\ref{sec:method}, we describe the data pre-processing and feature extraction procedures and the neural network models used to classify behavior from acoustic and lexical speech patterns.
The experimental setup is described in Sec.~\ref{sec:exp}, after which the behavior classification results are discussed and analyzed in Sec.~\ref{sec:results}.
Finally, we conclude in Sec.~\ref{sec:conclude} with some thoughts on possible extensions to this work.

\vspace{-0.5cm}

\begin{center}
\begin{table*}[ht]
\centering
\caption{Description of RMICS2 Behavior Codes}
\label{tab:rmics2}
\scalebox{0.88}{
\begin{tabular}{| p{1.7cm} | p{7cm} | p{7.3cm} |}
\hline
\textbf{Code} & \textbf{Definition} & \textbf{Examples (utterances not within context)} \\ \hline
High Hostile & Intense negative affect. Profoundly negative statements. Contempt, belligerence, character assassination. & \dq{Why are we still even in this relationship?} \dq{Of course you always think about yourself. That's just what you do.} \\ \hline
Low Hostile & Mild to medium intensity negative affect and verbal content that is mildly to moderately negative. Blame/ criticism (focus on behavior), demands. & \dq{I'm bothered by you not fixing anything around the house.} \dq{Shut up.} \\ \hline
Constructive Problem Discussion & All constructive approaches to discussing or solving problems. Includes descriptions of the problem, solutions, and questions. & \dq{When I take that medication I feel groggy.} \dq{We could be better at scheduling.} \\ \hline
Low Positive & Measured positive affect and statements with positive content (focused on others’ behavior) that facilitate low-level bonding within the couple. & \dq{I truly wish that we could be closer to one another} \dq{I appreciate the fact that you cleaned the house the other day without me asking you to do so.} \\ \hline
High Positive & Intense positive affect, statements with positive content that facilitate high-level bonding within the couple. & \dq{I love you} \dq{You're the funniest person I know.} \\ \hline
Dysphoric Affect & Sad or depressed expressed emotional states. & \dq{I'm useless. I can't even mow the lawn anymore.} \dq{I just don't know what else I can do.} \\ \hline
Other & Talk about the experimental situation. Not relevant to ongoing discussion. & \dq{Do you think the recorder stopped?} \\ \hline
\end{tabular}

}
\end{table*}
\end{center}

\section{Cancer Couples Interaction Dataset}
\label{sec:data}

\subsection{Data Collection}
\label{ssec:collection}

Data were gathered as part of a prospective observational study of couples coping with advanced cancer.
All procedures were conducted with the approval of the Institutional Review Board.
Advanced cancer patients and their spouse caregivers were recruited from thoracic and gastrointestinal clinics at a National Cancer Institute-designated Comprehensive Cancer Center.
Inclusion criteria for patients were (a) a diagnosis of stage III or IV non-small cell lung or pancreatic, esophageal, gastric, gallbladder, colorectal, hepatocellular, and bile duct cancers; (b) Karnofsky Performance Status score of 70+\footnote{
The Karnofsky Performance Scale Index allows patients to be classified as to their functional impairment. In short, above 70 patients are still able to mostly take care of themselves.}; (c) a prognosis of more than 6 months; and (d) undergoing active treatment at an NCI-designated Comprehensive Cancer Center.
Patients were to be cohabiting with a spouse/partner who self-identified as providing some care and also agreed to participate.
Participants were required to be over 18 years of age and English-speaking/writing.
A detailed description of study methods can be found elsewhere \cite{reblin2018everyday, reblin2018behind}.

Couples were asked to interact with each other in two structured discussions used in previous research \cite{manne2004couples}. 
First, couples engaged in a 10-minute neutral structured discussion (describing daily routines) which served as a baseline.
Next, participants independently completed the Cancer Inventory of Problem Situations, \cite{heinrich1984living} in which a list of 20 common cancer concerns (e.g. lack of energy, finances, over-protection) are rated as being not a problem, somewhat of a problem, or a severe problem.
After completing the concern list, items for which at least one person rated as a severe problem or both listed as at least somewhat of a problem were used as a prompt for the second structured discussion.
Couples were asked to have a 10-minute conversation in which they were asked to describe the issue, how it made them feel, and why they felt it was a concern.
In both interactions, the experimenter was present, but did not facilitate or participate in discussion.
Interactions were audio-recorded at 44.1 KHz in naturalistic environments (\eg clinic consult rooms, participant homes) using Sony Digital Recorders (ICD-UX533) with a lavalier microphone (Olympus ME-52W) worn by each participant.
After the structured discussions, couples were asked to continue wearing the audio recorders for the rest of the day to capture their natural interactions.

\subsection{Behavior Annotation}
\label{ssec:rmics2}

Two trained coders used Noldus Observer \cite{noldus1991observer} to review recordings and identify and timestamp communication behavior in the structured discussions using the Rapid Marital Interaction Coding System, 2nd Edition (RMICS2) \cite{heyman2004rapid, heyman1995marital}.
In RMICS2, the unit of analysis is speaker turn (i.e. each time an individual takes the floor to speak is a turn; an interaction where patient speaks, spouse speaks, and patient speaks again consists of 3 turns), and each speaker turn is labeled with a single hierarchical communication code.
A random sample of 20 percent of recordings was coded by both raters to calculate reliability.
Inter-rater agreement was excellent, with Kappas above .88 for all codes. 

The RMICS2 codes represent the emotional valence of the turn and are organized in a hierarchy (i.e., negative codes, positive codes, neutral code).
In addition to hostile, positive, and dysphoric affect codes, a constructive problem solving code represents a more emotionally-neutral discussion of the problem or conversational topic.
Table.~\ref{tab:rmics2} provides the definition of these codes along with representative examples.

\section{Methodology}
\label{sec:method}

\subsection{Data pre-processing}
\label{ssec:processing}

\subsubsection{Speech-Text Alignment}
\label{sssec:align}

For each session's audio recording, we had manual annotations of speaker segment time boundaries and behaviors, indicating the speaker label and behavior label of each audio segment.
The corresponding session transcript, however, contained only speaker turn text but no corresponding time boundaries or behavior labels.
Since training a lexical behavior classification system would require text samples labeled with behavior, we performed alignment between the audio and text using the gentle forced aligner \cite{gentle} tool.
Using the timing information of the aligned words in conjunction with the speaker time boundaries, we identified the text corresponding to each audio segment and, thus, obtained labeled text samples.

Visual inspection of the manual annotations revealed that the original speaker time boundaries were sometimes incorrectly marked, one reason for which could be the annotator's reaction time-lag \cite{mariooryad2013analysis}.
This would cause either a part of the speaker's audio and text to be lost or one speaker's audio and text to leak into the other's, therefore corrupting the data sample.
In order to rectify this, we leveraged the speaker turn label information from the transcript to accordingly adjust the time boundaries forward or backward, depending on the shift in the annotation.
The resulting new time boundaries were used as ground-truth information in our experiments.

\subsubsection{Labels}
\label{sssec:labels}

The ordinal set of RMICS2 behaviors consists of 5 codes: \dq{High Hostile}, \dq{Low Hostile}, \dq{Constructive Problem Discussion}, \dq{Low Positive} and \dq{High Positive}, in this order.
However, since the number of samples in all classes except \dq{Constructive Problem Discussion} was small, we combined the \dq{High} and \dq{Low} codes into a single class for both Positive and Hostile.
This resulted in 3 behavior classes : \dq{Hostile}, \dq{Constructive} and \dq{Positive}.
In this work, we focus only on the ordinal set of behaviors; hence, we did not use the remaining independent codes \dq{Dysphoric Affect} and \dq{Other}.

\subsubsection{Partitions}
\label{sssec:partitions}

Speaker attributes such as gender have been shown to be related to aspects of behavior expression and perception \cite{tseng2018honey} and emotion recognition \cite{alaerts2011action}.
The topic of conversation (\eg   sensitive issues) can also cause some behaviors to be expressed more strongly than others.
Thus, in addition to classifying behavior based on turn-level speech and text, we also investigate the effect of conditioning this task on different attributes of the speaker and interaction.
We do so by partitioning the data based on the attribute, training and testing models for each partition separately and comparing their performance to that of the models trained on the original, un-partitioned data.
In this paper, we focus on three partitions:

\begin{enumerate}[itemsep=0em]
    \item Speaker Gender: Male / Female
    \item Speaker Role: Patient / Caregiver
    \item Interaction Content: Neutral / Stress
\end{enumerate}

In the \dq{Role} partition, \emph{Patient} refers to the speaker recruited from the clinic and \emph{Caregiver} refers to their spouse.
In the \dq{Content} partition, as mentioned in Sec.~\ref{ssec:collection}, \emph{Neutral} refers to the part of the interaction where speakers discussed their daily routines while \emph{Stress} refers to the part where they discussed a cancer concern.
Table.~\ref{tab:data_parts} shows the number of samples per behavior for each partition after pre-processing.

\begin{table}[]
\caption{Number of samples per behavior for different partitions}
\label{tab:data_parts}
\begin{tabular}{c p{0.9cm} c p{0.9cm} c p{1cm} c p{1cm} c p{1cm}}
\hline
\textbf{Partition} & & \begin{tabular}[c]{@{}c@{}}\textbf{Constructive} \end{tabular} & \textbf{Hostile} & \textbf{Positive} \\ \hline
 None &             & 13450             & 176          &1369            \\ \hline
\multirow{2}{*}{Gender}  & Male      &6673              & 72          & 715           \\ \cline{2-2}
                         & Female    &6777              & 104          &  654          \\ \hline
\multirow{2}{*}{Role}    & Patient   & 6670             & 76          & 728           \\ \cline{2-2} 
                         & Caregiver & 6780             & 100          & 641           \\ \hline
\multirow{2}{*}{Content} & Neutral   & 7584             & 54          & 467           \\ \cline{2-2} 
                         & Stress    & 5866             & 122          &   902         \\ \hline
\end{tabular}
    \vspace{-1ex}
\end{table}

\subsection{Features}
\label{ssec:features}
\subsubsection{Acoustic}
For acoustic modality, we extract features using openSMILE toolkit \cite{eyben2013recent} with the default eGeMAPS configuration.
The eGeMAPS feature set includes 88 features related to frequency, energy, spectral, cepstral, and dynamic information. 
The final feature vectors of each utterance are generated by calculating the following statistics: mean, coefficient of variation, 20th, 50th, and 80th percentile, range of 20th to 80th percentile, the spectral slopes from 0-500Hz and 500-1500Hz, mean of the Alpha Ratio, the Hammarberg Index, and mean and standard deviation of the slope of rising/falling signal parts.

\subsubsection{Lexical}
We use sentence representations as lexical features since we want to classify behaviors based on the text of an utterance turn.
In this work, we use the 600-dimensional sentence embeddings proposed by Tseng et al. \cite{tseng2018multi} where they were shown to perform better on affective tasks than other general purpose embeddings.
These embeddings were obtained from a model trained using multitask learning of predicting dialogue turn sequences from movies and sentiment classification, both of which are complementarily matched to our domain.
\subsection{Models}
\label{ssec:models}


For both the acoustic and lexical modalities, we used Feed-Forward Deep Neural Networks (DNNs) to classify behavior labels based on turn-level features.
The acoustic system DNN has 3 to 4 hidden layers while the lexical system has 1 to 3 hidden layers and both use ReLU activation functions. We tried all hyperparameter combinatorics shown in Table.~\ref{tab:model_config}.

\begin{table}[t]
\caption{Model hyperparameters}
\label{tab:model_config}
\scalebox{0.95}{
\begin{tabular}{c l l p{3cm}}
\hline
\multirow{2}{*}{Acoustic} & \multicolumn{1}{c}{\begin{tabular}[c]{@{}c@{}}hidden layers\\ configuration\end{tabular}} & \begin{tabular}[c]{@{}l@{}}(64, 32, 16), (128, 64, 32), \\ (128, 64, 32, 32)\end{tabular}   \\ \cline{2-2} 
                          & batch size    & 32, 64, 128, 256 \\  \cline{2-2}
                          & class weights    & $1-\frac{x_i}{\sum_{i}(x_i)}$, $\frac{\sum_{i}(x_i)}{x_i}$, $\frac{\max_{i}x_i}{x_i}$ \\ \hline
\multirow{2}{*}{Lexical} & \multicolumn{1}{c}{\begin{tabular}[c]{@{}c@{}}hidden layers\\ configuration\end{tabular}} & \begin{tabular}[c]{@{}l@{}}(300, 200, 100), (200, 100, 50),\\ (300, 200), (200, 100), (100, 50),\\ (300), (200), (100), (50)\end{tabular}\\ \cline{2-2} 
                          & batch size    & 25                                 \\ \cline{2-2} 
                          & rate decay  & 1e-1, 5e-1                          \\  \cline{2-2} 
                          & class weights    & $\frac{\max_{i}x_i}{x_i}$ \\ \hline
\multirow{2}{*}{Training} & optimizer & Adam, SGD \\ \cline{2-2}
 & learning rate & 1e-2, 1e-3, 1e-4 \\ \hline                            
\end{tabular}
}
\end{table}

\subsection{Sample Weighing}
To address the high class imbalance, during training, we employ weighted categorical cross entropy loss.
Assuming that the number of samples in each class is $x_i$, where $i \in [1, 2, 3]$ in our case, we use three methods to calculate the class weight $w_i$:

\vspace{-.1in}
\begin{table}[h]
\centering
\begin{tabular}{ccc}
(1) $1-\frac{x_i}{\sum_{i}(x_i)}$ & (2) $\frac{\sum_{i}(x_i)}{x_i}$ & (3) $\frac{\max_{i}x_i}{x_i}$
\end{tabular}
\end{table}
\vspace{-.15in}

When optimizing for the acoustic system, no major difference was observed between the weighing methods 2 \& 3, while method 1 was under performing. We thus only employed method 3 for the lexical system.

\section{Experimental Setup}
\label{sec:exp}

We use a leave-one-couple-out nested cross-validation paradigm for the behavior classification task.
In every test fold \emph{i}, we set the samples of couple $C_i$ as test data and use the samples of remaining couples $C_j, j \neq i$ to train and validate our models.
We then pick the model that performed best on the validation data, run it on the samples of couple $C_i$ and compute the test evaluation metric.
Since our data has 85 unique couples, this process is repeated over 85 folds and the test evaluation metric is averaged over all of them.

In every test fold, we create an 80\% training, 20\% validation split such that every couple appears in only one split and each split contains samples of all 3 behaviors.
For model evaluation and checkpoint selection, we use the Unweighted Average Recall (UAR) metric which accounts for class imbalance.
The model hyperparameters, such as number of hidden layers, hidden layer size, learning rate, etc. (shown in Table.~\ref{tab:model_config}), are tuned on the validation data using grid search.

In partition experiments, since one model is built for each partition, we halve the model size so that the total number of parameters remains the same as when not using any partitions.
We used \emph{PyTorch} \cite{paszke2017automatic} and \emph{scikit-learn} \cite{scikit-learn} in our experiments.

\section{Results \& Discussion}
\label{sec:results}

\subsection{Performance}
\label{ssec:perf}

\begin{table}[!t]
\centering
\caption{Mean (Std. Deviation) UAR \% of Test-fold Behavior Classification}
\label{tab:results}
\begin{tabular}{ccc}
\toprule
\textbf{Partition} & \multicolumn{2}{c}{\textbf{System}} \\
\midrule
 & Acoustic & Lexical \\
\midrule
None & 45.36 (12.85) & 57.42 (14.45) \\
Gender & 44.00 (12.79) & 56.85 (14.57) \\
Role & 42.38 (11.83) & 55.38 (13.84) \\
Content & 45.40 (14.49) & 56.40 (15.44) \\
\bottomrule
\end{tabular}
\end{table}

Table.~\ref{tab:results} shows the average UAR\%, computed across all test folds, for each modality and partitioning scheme.
The best performing classifier is the lexical system trained on the entire data without partitioning, which achieves 57.42\%.
For comparison, we computed the average UAR\% due to chance, by randomly assigning labels according to the class priors, which was found to be 43\%.
Our results demonstrate the potential feasibility of using a system to automatically identify these behaviors based on conversational speech cues.

Comparing modalities, we see that the lexical system performs better than the acoustic system across all partitioning schemes.
This is consistent with findings from past relevant work \cite{tseng2018honey,black2011you} and is likely due to the fact that the lexical features are computed on human transcriptions while the acoustic features are computed on the noisy raw audio signal.

We also tested feature-level and decision-level fusion of the acoustic and lexical modalities but they did not perform better than the lexical modality.
An inspection of the best hyperparameters from every test fold revealed a trend towards larger models, Adam optimizer and 1e-3 learning rate for the lexical system and larger batch sizes and ratio-based class weights for the acoustic system.

Comparing partitions, we see that conditioning on the content of the interaction works best for the acoustic modality whereas using the data without any partitions works best for lexical.
It is also observed that accounting for the speaker's gender did not improve the result for both lexical and acoustic systems, implying that there might not exist a significant gender difference in how spouses express these behaviors. This matches our domain-experts' experience and expectation.

\subsection{Analysis}
\label{ssec:analysis}

While inspecting our model predictions, we observed that utterances neighboring the occurrences of \dq{Hostile} and \dq{Positive} were also labeled as such in many cases.
A possible explanation for this phenomenon is the presence of annotation delay attributed to the annotators' reaction lag \cite{mariooryad2013analysis}, wherein annotations are positioned multiple seconds after the actual occurrence of the behavior.
We refer to this as a delayed, or \dq{backward} mislabeling of the ground-truth.
Conversely, labeling an occurrence as hostile or positive might lead an annotator to disregard residual but similar behavior in subsequent occurrences, resulting in a \dq{forward} mislabeling of the ground-truth.

Further to the mislabeling possibility, the ability to coarsely identify a region of hostility is more important than identifying the specific utterance labeled as such.
Hostility usually builds up and human experts before any manual inspection would still require context to better understand the issue.

\begin{figure}[tb!]
    \centering
    \includegraphics[width=0.8\linewidth]{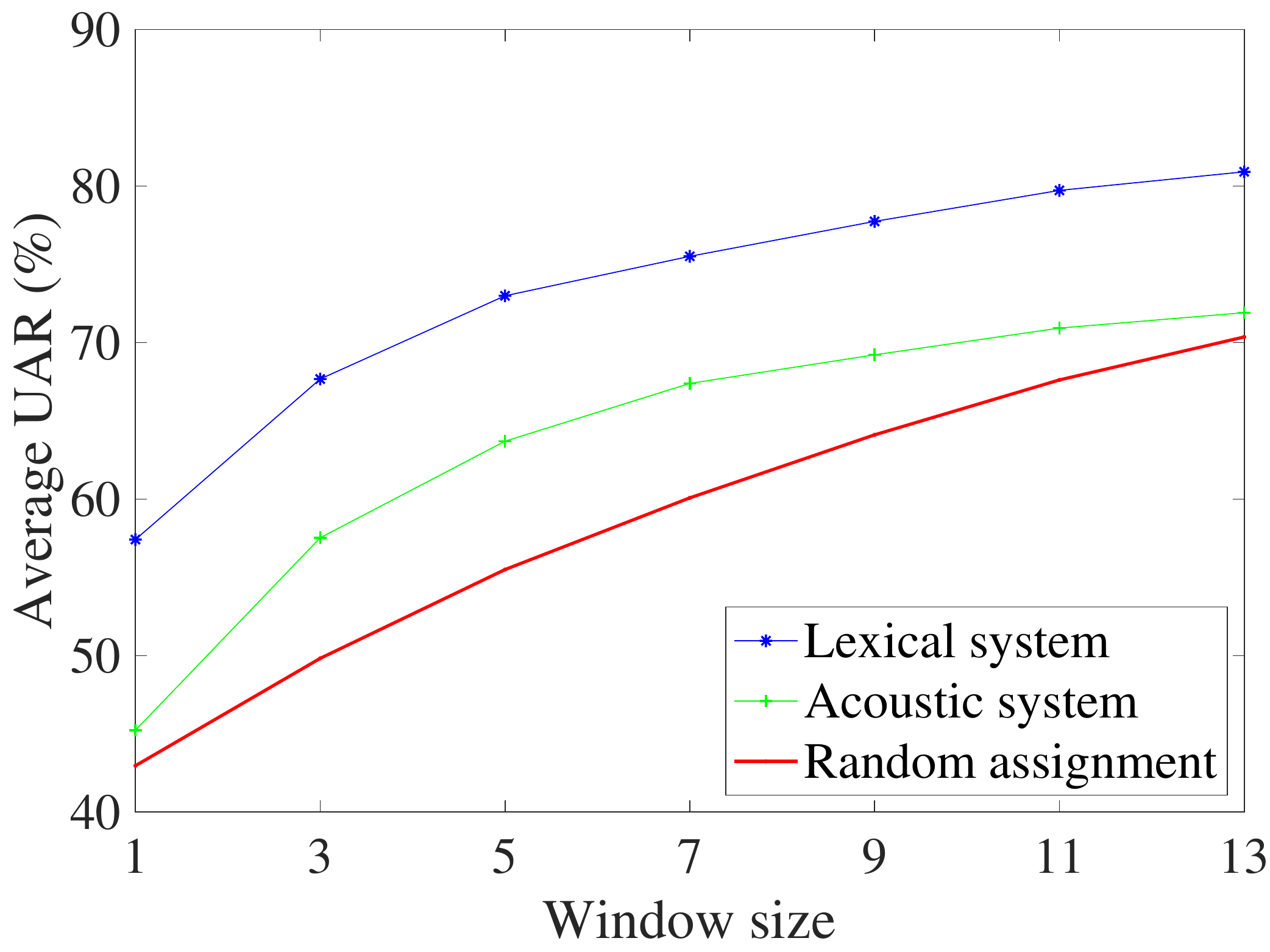}
    \caption{Test recall versus extended evaluation window size}
    \vspace{-2ex}
    \label{fig:label_smoothing}
    \vspace{-3ex}
\end{figure}

We thus tested identifying for the presence of \dq{Hostile} and \dq{Positive} within a \dq{tolerance} window, where a classification was judged to be correct if it was present inside a symmetric window centered on the target utterance.
Test results using this evaluation are shown in Figure \ref{fig:label_smoothing}, where window size \emph{K}+1 represents tolerance of $\frac{K}{2}$ neighboring utterances on each side.

We observed that the average UAR\% for both acoustic and lexical systems increases at a higher rate than that for chance when using a window size of 3.
Moreover, 54.35\% of the false negatives in utterances whose ground truth label was either \dq{Hostile} or \dq{Positive} are corrected when using a window size of 5.
These findings offer potential empirical evidence of the need to often consider multiple turns to identify such behaviors and that there may be some annotation offset inherent to the way humans integrate information.
Furthermore, they encourage the use of a context-based approach, where the neighboring turns of an utterance are also considered when classifying its behavior.

\vspace{-0.5ex}
\section{Conclusions \& Future Work}
\label{sec:conclude}
\vspace{-0.5ex}
In this paper, we introduced a dataset of real-life interactions between couples where one of the spouses has been diagnosed with cancer.
We also described neural network-based systems for classifying the spouses' behaviors based on the language and vocal characteristics they use when conversing with each other.
Our results showed that the lexical system performed better than the acoustic system and that there were no major differences between different genders, patient/caregiver roles and interaction content with respect to how spouses expressed behaviors.
A windowed-evaluation analysis of our results indicated that affective behavior might manifest in a region around target instances and, thus, context-based classification approaches might be more effective at identifying them.

In future work, we will employ automatic transcription for the lexical modality and investigate advanced methods for fusing acoustic and lexical systems.
We also plan on exploring data augmentation techniques to deal with the problem of inadequate number of samples for some behavior classes.
We will also investigate the ordinal and symmetric relationships between these behaviors towards more accurate classifications.

Critically and importantly, we will implement an iterative human-machine processing pipeline to enable rapid transcription and coding of newly collected data, thus expanding the understanding of cancer communication.
\vspace{-0.5ex}
\section{Acknowledgements}

Funding was provided by the American Cancer Society ACS MRSG 13-234-01-PCSM; PI: Reblin.
This research was also supported in part by NSF, NIH, and DOD.
\newpage

\bibliographystyle{IEEEtran}
\bibliography{mybib}

\begin{thebibliography}{10}
\providecommand{\url}[1]{#1}
\csname url@samestyle\endcsname
\providecommand{\newblock}{\relax}
\providecommand{\bibinfo}[2]{#2}
\providecommand{\BIBentrySTDinterwordspacing}{\spaceskip=0pt\relax}
\providecommand{\BIBentryALTinterwordstretchfactor}{4}
\providecommand{\BIBentryALTinterwordspacing}{\spaceskip=\fontdimen2\font plus
\BIBentryALTinterwordstretchfactor\fontdimen3\font minus
  \fontdimen4\font\relax}
\providecommand{\BIBforeignlanguage}[2]{{%
\expandafter\ifx\csname l@#1\endcsname\relax
\typeout{** WARNING: IEEEtran.bst: No hyphenation pattern has been}%
\typeout{** loaded for the language `#1'. Using the pattern for}%
\typeout{** the default language instead.}%
\else
\language=\csname l@#1\endcsname
\fi
#2}}
\providecommand{\BIBdecl}{\relax}
\BIBdecl

\bibitem{cancer2019report}
{\relax American Cancer Society}, ``Cancer facts \& figures 2019.'' 2019.

\bibitem{national2016caregiving}
{\relax National Alliance for Caregiving in partnership with the National
  Cancer Institute Cancer Support Community}, ``Cancer caregiving in the {U.S.}
  -- an intense, episodic, and challenging care experience,'' 2016.

\bibitem{li2014literature}
Q.~Li and A.~Y. Loke, ``A literature review on the mutual impact of the spousal
  caregiver–cancer patients dyads: ‘communication’, ‘reciprocal
  influence’, and ‘caregiver–patient congruence’,'' \emph{European
  Journal of Oncology Nursing}, vol.~18, no.~1, pp. 58 -- 65, 2014.

\bibitem{traa2015dyadic}
M.~J. Traa, J.~De~Vries, G.~Bodenmann, and B.~L. Den~Oudsten, ``Dyadic coping
  and relationship functioning in couples coping with cancer: a systematic
  review,'' \emph{British Journal of Health Psychology}, vol.~20, no.~1, pp.
  85--114, 2015.

\bibitem{badr2017new}
H.~Badr, ``New frontiers in couple-based interventions in cancer care: refining
  the prescription for spousal communication,'' \emph{Acta Oncologica},
  vol.~56, no.~2, pp. 139--145, 2017.

\bibitem{reblin2018everyday}
M.~Reblin, R.~E. Heyman, L.~Ellington, B.~R. Baucom, P.~G. Georgiou, and S.~T.
  Vadaparampil, ``Everyday couples’ communication research: Overcoming
  methodological barriers with technology,'' \emph{Patient education and
  counseling}, vol. 101, no.~3, pp. 551--556, 2018.

\bibitem{black2013toward}
M.~P. Black, A.~Katsamanis, B.~R. Baucom, C.-C. Lee, A.~C. Lammert,
  A.~Christensen, P.~G. Georgiou, and S.~S. Narayanan, ``Toward automating a
  human behavioral coding system for married couples’ interactions using
  speech acoustic features,'' \emph{Speech communication}, vol.~55, no.~1, pp.
  1--21, 2013.

\bibitem{li2016_sparsely-connec}
H.~Li, B.~Baucom, and P.~Georgiou, ``Sparsely connected and disjointly trained
  deep neural networks for low resource behavioral annotation: Acoustic
  classification in couples’ therapy,'' \emph{Interspeech 2016}, pp.
  1407--1411, 2016.

\bibitem{tseng2016couples}
S.-Y. Tseng, S.~N. Chakravarthula, B.~R. Baucom, and P.~G. Georgiou, ``Couples
  behavior modeling and annotation using low-resource lstm language models.''
  in \emph{INTERSPEECH}, 2016, pp. 898--902.

\bibitem{gupta2014multimodal}
R.~Gupta, N.~Malandrakis, B.~Xiao, T.~Guha, M.~Van~Segbroeck, M.~Black,
  A.~Potamianos, and S.~Narayanan, ``Multimodal prediction of affective
  dimensions and depression in human-computer interactions,'' in
  \emph{Proceedings of the 4th International Workshop on Audio/Visual Emotion
  Challenge}.\hskip 1em plus 0.5em minus 0.4em\relax ACM, 2014, pp. 33--40.

\bibitem{nasir2016multimodal}
M.~Nasir, A.~Jati, P.~G. Shivakumar, S.~Nallan~Chakravarthula, and P.~Georgiou,
  ``Multimodal and multiresolution depression detection from speech and facial
  landmark features,'' in \emph{Proceedings of the 6th International Workshop
  on Audio/Visual Emotion Challenge}.\hskip 1em plus 0.5em minus 0.4em\relax
  ACM, 2016, pp. 43--50.

\bibitem{tanaka2017brain}
T.~Tanaka, T.~Yamamoto, and M.~Haruno, ``Brain response patterns to economic
  inequity predict present and future depression indices,'' \emph{Nature Human
  Behaviour}, vol.~1, no.~10, p. 748, 2017.

\bibitem{cummins2015review}
N.~Cummins, S.~Scherer, J.~Krajewski, S.~Schnieder, J.~Epps, and T.~F.
  Quatieri, ``A review of depression and suicide risk assessment using speech
  analysis,'' \emph{Speech Communication}, vol.~71, pp. 10--49, 2015.

\bibitem{venek2017adolescent}
V.~Venek, S.~Scherer, L.-P. Morency, J.~Pestian \emph{et~al.}, ``Adolescent
  suicidal risk assessment in clinician-patient interaction,'' \emph{IEEE
  Transactions on Affective Computing}, vol.~8, no.~2, pp. 204--215, 2017.

\bibitem{baucom2017promise}
B.~Baucom, P.~Georgiou, C.~Bryan, E.~Garland, F.~Leifker, A.~May, A.~Wong, and
  S.~Narayanan, ``The promise and the challenge of technology-facilitated
  methods for assessing behavioral and cognitive markers of risk for suicide
  among us army national guard personnel,'' \emph{International journal of
  environmental research and public health}, vol.~14, no.~4, p. 361, 2017.

\bibitem{nasir2017complexity}
M.~Nasir, B.~R. Baucom, C.~J. Bryan, S.~S. Narayanan, and P.~G. Georgiou,
  ``Complexity in speech and its relation to emotional bond in
  therapist-patient interactions during suicide risk assessment interviews.''
  in \emph{INTERSPEECH}, 2017, pp. 3296--3300.

\bibitem{reblin2018behind}
M.~Reblin, S.~K. Sutton, S.~T. Vadaparampil, R.~E. Heyman, and L.~Ellington,
  ``Behind closed doors: How advanced cancer couples communicate at home,''
  \emph{Journal of psychosocial oncology}, pp. 1--14, 2018.

\bibitem{manne2004couples}
S.~Manne, M.~Sherman, S.~Ross, J.~Ostroff, R.~E. Heyman, and K.~Fox, ``Couples'
  support-related communication, psychological distress, and relationship
  satisfaction among women with early stage breast cancer.'' \emph{Journal of
  consulting and clinical psychology}, vol.~72, no.~4, p. 660, 2004.

\bibitem{heinrich1984living}
R.~L. Heinrich, C.~C. Schag, and P.~A. Ganz, ``Living with cancer: The cancer
  inventory of problem situations,'' \emph{Journal of Clinical Psychology},
  vol.~40, no.~4, pp. 972--980, 1984.

\bibitem{noldus1991observer}
L.~P. Noldus, ``The observer: a software system for collection and analysis of
  observational data,'' \emph{Behavior Research Methods, Instruments, \&
  Computers}, vol.~23, no.~3, pp. 415--429, 1991.

\bibitem{heyman2004rapid}
R.~E. Heyman, ``Rapid marital interaction coding system (rmics),'' in
  \emph{Couple observational coding systems}.\hskip 1em plus 0.5em minus
  0.4em\relax Routledge, 2004, pp. 81--108.

\bibitem{heyman1995marital}
R.~E. Heyman, R.~L. Weiss, and J.~M. Eddy, ``Marital interaction coding system:
  Revision and empirical evaluation,'' \emph{Behaviour Research and Therapy},
  vol.~33, no.~6, pp. 737--746, 1995.

\bibitem{gentle}
``gentle forced aligner,'' \url{https://github.com/lowerquality/gentle}.

\bibitem{mariooryad2013analysis}
S.~Mariooryad and C.~Busso, ``Analysis and compensation of the reaction lag of
  evaluators in continuous emotional annotations,'' in \emph{2013 Humaine
  Association Conference on Affective Computing and Intelligent
  Interaction}.\hskip 1em plus 0.5em minus 0.4em\relax IEEE, 2013, pp. 85--90.

\bibitem{tseng2018honey}
S.-Y. Tseng, H.~Li, B.~Baucom, and P.~Georgiou, ``Honey, i learned to talk:
  Multimodal fusion for behavior analysis,'' in \emph{Proceedings of the 2018
  on International Conference on Multimodal Interaction}.\hskip 1em plus 0.5em
  minus 0.4em\relax ACM, 2018, pp. 239--243.

\bibitem{alaerts2011action}
K.~Alaerts, E.~Nackaerts, P.~Meyns, S.~P. Swinnen, and N.~Wenderoth, ``Action
  and emotion recognition from point light displays: an investigation of gender
  differences,'' \emph{PloS one}, vol.~6, no.~6, p. e20989, 2011.

\bibitem{eyben2013recent}
F.~Eyben, F.~Weninger, F.~Gross, and B.~Schuller, ``Recent developments in
  opensmile, the munich open-source multimedia feature extractor,'' in
  \emph{Proceedings of the 21st ACM international conference on
  Multimedia}.\hskip 1em plus 0.5em minus 0.4em\relax ACM, 2013, pp. 835--838.

\bibitem{tseng2018multi}
S.-Y. Tseng, B.~Baucom, and P.~Georgiou, ``Unsupervised online multitask
  learning of behavioral sentence embeddings,'' \emph{arXiv preprint
  arXiv:1807.06792}, 2018.

\bibitem{paszke2017automatic}
A.~Paszke, S.~Gross, S.~Chintala, G.~Chanan, E.~Yang, Z.~DeVito, Z.~Lin,
  A.~Desmaison, L.~Antiga, and A.~Lerer, ``Automatic differentiation in
  pytorch,'' \emph{NIPS 2017 workshop}, 2017.

\bibitem{scikit-learn}
F.~Pedregosa, G.~Varoquaux, A.~Gramfort, V.~Michel, B.~Thirion, O.~Grisel,
  M.~Blondel, P.~Prettenhofer, R.~Weiss, V.~Dubourg, J.~Vanderplas, A.~Passos,
  D.~Cournapeau, M.~Brucher, M.~Perrot, and E.~Duchesnay, ``Scikit-learn:
  Machine learning in {P}ython,'' \emph{Journal of Machine Learning Research},
  vol.~12, pp. 2825--2830, 2011.

\bibitem{black2011you}
M.~P. Black, P.~G. Georgiou, A.~Katsamanis, B.~R. Baucom, and S.~Narayanan,
  ``“you made me do it”: Classification of blame in married couples'
  interactions by fusing automatically derived speech and language
  information,'' in \emph{Twelfth Annual Conference of the International Speech
  Communication Association}, 2011.

\end{thebibliography}

\end{document}